\documentclass[runningheads]{llncs}
\usepackage{graphicx}

\usepackage{times}
\usepackage{soul}
\usepackage{url}
\usepackage[hidelinks]{hyperref}
\usepackage[utf8]{inputenc}
\usepackage[small]{caption}
\usepackage{graphicx}
\usepackage{amsmath}
\usepackage{booktabs}
\usepackage{algorithm}
\usepackage{algorithmic}
\usepackage{color}
\usepackage{multirow}
\usepackage{anyfontsize}
\usepackage{subfig,wrapfig}

\makeatletter
\newcommand{\printfnsymbol}[1]{%
  \textsuperscript{\@fnsymbol{#1}}%
}

\newcommand{\calC}{\mathcal{C}}
\newcommand{\calX}{\mathcal{X}}
\newcommand{\calL}{\mathcal{L}}

\begin{document}
\title{Primate Face Identification in the Wild}
%
%
\author{\thanks{Equal Contribution}Ankita Shukla\inst{1}\orcidID{0000-0002-1878-2667} \and
\printfnsymbol{1}Gullal Singh Cheema \inst{1}\orcidID{0000-0003-4354-9629} \and
Saket Anand\inst{1}\and Qamar Qureshi\inst{2}\and  Yadvendradev Jhala \inst{2}}
\authorrunning{A. Shukla et al.}
%
\institute{Indraprastha Institute of Information Technology Delhi, India \and
Wildlife Institute of India, Dehradun, India
\\
\email{\{ankitas,gullal1408,anands\}@iiitd.ac.in, \{qnq,jhalay
\}@wii.gov.in}}
\maketitle              
\begin{abstract}
Ecological imbalance owing to rapid urbanization and deforestation has adversely affected the population of several wild animals. This loss of habitat has skewed the population of several non-human primate species like chimpanzees and macaques and has constrained them to co-exist in close proximity of human settlements, often leading to human-wildlife conflicts while competing for resources. For effective wildlife conservation and conflict management, regular monitoring of population and of conflicted regions is necessary. However, existing approaches like field visits for data collection and manual analysis by experts is resource intensive, tedious and time consuming, thus necessitating an automated, non-invasive, more efficient alternative like image based facial recognition. The challenge in individual identification arises due to unrelated factors like pose, lighting variations and occlusions due to the uncontrolled environments, that is further exacerbated by limited training data. Inspired by human perception, we propose to learn representations that are robust to such nuisance factors and capture the notion of similarity over the individual identity sub-manifolds. The proposed approach, Primate Face Identification (PFID), achieves this by training the network to distinguish between positive and negative pairs of images. The PFID loss augments the standard cross entropy loss with a pairwise loss to learn more discriminative and generalizable features, thus making it appropriate for other related identification tasks like open-set, closed set and verification. We report state-of-the-art accuracy on facial recognition of two primate species, rhesus macaques and chimpanzees under the four protocols of classification, verification, closed-set identification and open-set recognition.
\keywords{Face Recognition  \and Deep Learning \and Primates \and Social Good}
\end{abstract}
\section{Introduction}
One of the key indicators of a healthy ecosystem is its constituent biodiversity. Over the last several decades, technological progress has substantially improved human quality of life, albeit at a cost of rapid environmental degradation. Specifically, to meet the needs of the growing human population, various factors like urban and infrastructural development, agricultural land expansion and livestock ranching have resulted in soaring rates of deforestation. In addition to the risk of extinction for many species, shrinking natural habitats have led to increased interactions between humans and wildlife, often raising safety concerns for both.  

Conflicts with primarily forest-dwelling species like big cats (tigers, leopards, mountain lions, etc.), elephants, bears or wolves may cause severe injuries or even death to humans. On the other hand, there are species which have transitioned into a commensal relationship with humans, i.e., they rely on humans for food without causing direct harm. Due to their apparent harmlessness, several commensal (or semi-commensal) species like wild herbivores, wild boars, macaques and other non-human primates often dwell in close proximity of human settlements. This co-existence leads to indirect conflicts in the form of crop-raiding and property damage as well as occasional direct conflicts such as attacks or biting incidents. An example image of crop raiding and primates in close vicinity of humans is shown in Figure \ref{fig: intro_example}. Certain species like the rhesus macaque (\emph{Macaca mulatta}) have become a cause of serious concern due to their resilience and ability to co-exist with humans in rural, semi-urban and urban areas.  Their prolific breeding and short gestation periods lead to high population densities, thereby increasing the chances and extent of conflicts with humans. 
\vspace{-0.4cm}
\begin{figure}[h]
	\centering
			{
			\includegraphics[width =0.42\textwidth]{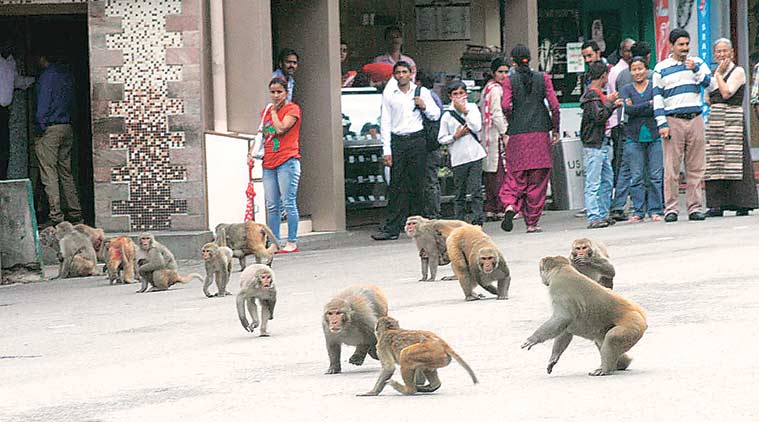}
        \includegraphics[width  =0.42\textwidth]{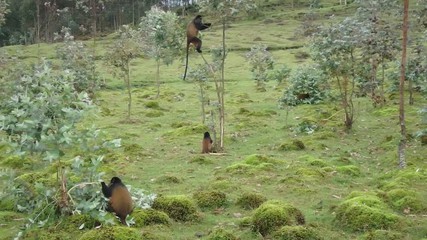}
        }
		\caption{Example images showing primates in human shared space and crop raiding [source: google images].}
		\label{fig: intro_example}
        \vspace{-0.5cm}	
\end{figure}
As a consequence, organizations have resorted to lethal conflict management measures like culling \cite{AndersonHistory_2016}, which become infeasible when the conflicted species have declining populations, e.g., the human-primate conflict crisis in Sri Lanka where two of the responsible primate species are endangered: Toque macaques (\emph{Macaca sinica}) and the purple faced langur (\emph{Trachypithecus vetulus}) \cite{Srilanka_2018}. Besides, the effectiveness of lethal measures is well debated and poorly designed initiatives could have unexpected consequences like increased aggression or even extinction of the conflicted species \cite{Nyhus2016}. On the other hand, non-lethal approaches are easier to adopt across geographies as they avoid complex socio-religious issues \cite{Saraswat2015}. Two recurring non-lethal themes in conflict management discussions are population monitoring and stakeholder engagement \cite{Nyhus2016}, both of which can be easily achieved with a combination of smartphone and AI technology. Pursuing a crowdsourcing approach to population monitoring and conflict reporting has two direct benefits: the cost and scalability of data collection for population monitoring can be improved drastically and active involvement of the affected community can help increase awareness, which in turn abates the human behavioral factors that often influence human-wildlife conflicts. 

In this work, we focus on addressing the human-primate conflicts, largely because of the frequency and magnitude of encounters in urban, rural and agricultural regions across developing South Asian nations \cite{Anand2017}. Inspired by the success and scalability of human face recognition, we propose a Primate Face Identification (PFID) system. Automatic identification capabilities could serve as a backbone for a crowdsourcing platform, where geo-referenced images submitted by users are automatically indexed by individuals, gender, age, etc. Such an indexed database could simplify downstream tasks like primate population monitoring and analysis of conflict reports, enabling better informed and effective strategies for conflict as well as conservation management.  
We summarize the contributions of this paper as follows:
\begin{itemize}
    \item We propose \emph{Primate Face Identification} (PFID), a deep neural network based system for automated identification of individual primates using facial images.
    \item We introduce a guided pairwise loss using similar and dissimilar image pairs to learn robust and generalizable representations.
    \item Our fully automatic pipeline convincingly beats state-of-the-art methods on two datasets (macaques and chimpanzees) under \emph{all} settings.
\end{itemize}

\section{Existing Work on Face Recognition}
There is a vast body of literature in human face recognition. Without attempting to present a comprehensive survey, we briefly discuss prior work relevant to facial identification of primates. We broadly categorize these approaches into two categories: Non Deep Learning Approaches and Deep Learning Approaches. 
\\\\
\noindent{\textbf{Non Deep Learning Approaches}}
Traditional face recognition pipelines comprised of face alignment, followed by low level feature extraction and classification. Early works in primate face recognition \cite{Loos2012}, adapted the Randomfaces \cite{Robust_2009} technique for identifying chimpanzees in the wild and follows the standard pipeline for face recognition. Later, LemurID was proposed in \cite{LemurID_2017}, which additionally used manual marking of the eyes for face alignment. Patch-wise multi-scale Local Binary Pattern (LBP) features were extracted from aligned faces and used with LDA to construct a representation, which was then used with an appropriate similarity metric for identifying individuals. 
\\\\
\noindent{\textbf{Deep Learning Approaches}}
Freytag et al. \cite{freytag2016chimpanzee} used Convolutional Neural Networks (CNNs) for learning a feature representation of chimpanzee faces. For increased discriminative power, the architecture uses a bilinear pooling layer after the fully connected layers (or a convolutional layer), followed by a matrix log operation. These features are then used to train an SVM classifier for classification of known identities. Later, \cite{Brust_2017} developed face recognition for gorilla images captured in the wild. This approach fine-tuned a YOLO detector \cite{YOLO_CVPR2016} for gorilla faces. For classification, a similar approach was taken as \cite{freytag2016chimpanzee}, where pre-trained CNN features are used to train a linear SVM. More recently, \cite{Jain_2018} proposed PrimNet, a deep neural net based approach that uses the \emph{Additive Margin Softmax} loss \cite{Wang_AMSoftmax_2018} and achieves state of the art performance for identifying individuals across different primate species including lemur, chimpanzee and golden monkey. However, it requires substantial manual effort to designing landmark templates for face alignment prior to identification process, which can adversely affect adoption rates in a crowdsourced mobile app setting. For human face recognition techniques, various approaches have improved performance by combining the standard cross entropy loss with other loss functions such as contrastive loss \cite{Sun_NIPS2014} and center loss \cite{Wen_ECCV2016} to learn more discriminative features.



\section{Primate Face Identification (PFID) System}
\paragraph{Pose Invariant Representation Learning}
We would like to motivate the choice of our loss function with the following reasons
\begin{enumerate}
    \item Our approach is inspired by the human perception system, which is robust to nuisance factors like illumination and pose and is able to identify individual faces captured in unconstrained environments and extreme poses. Geometrically, face images of an identity defines a sub-manifold \cite{murphy2008head} in image manifold of faces. This allows one to devise a metric such that sample pairs of the same identity have small distances regardless of pose and other nuisance factors, while those of different identities have larger distances. In PFID, we use a deep neural network to learn such a representation through a specially designed loss function over similar and dissimilar pairs of primate face images.  
    
    \item  Learning invariant features has long been a challenging issue in computer vision. Owing to the high curvature of original image data manifold \cite{lu1998image} , simple metric like euclidean distance fails to capture the underlying data semantics. Consequently, linear methods also are inappropriate to learn decision boundaries for tasks like image recognition. In such scenarios, deep learning approaches have come in handy, with their ability to flatten the data manifold owing to the successive non linear operations applied though a series of layers \cite{Brahma_DLworks_2016}. However, deep models are often trained with a cross-entropy based classification loss, to drive the class probability distribution for a given image independently to one hot encoding vector. Given sufficient training data, this training protocol often generalizes well for classification task, however, its performance is often limited on other related tasks like verification and unseen class generalizabilty. The latter's performance crucially depends on the ability to learn a representation space that can model class-level similarities. By incorporating a pairwise similarity loss term operating on the class probability (softmax) distributions, we drive the learned representations to be semantically more meaningful, and hence invariant to other factors.
    \end{enumerate}
\begin{figure}[h!]
	\centering
			\includegraphics[width =0.85\textwidth]{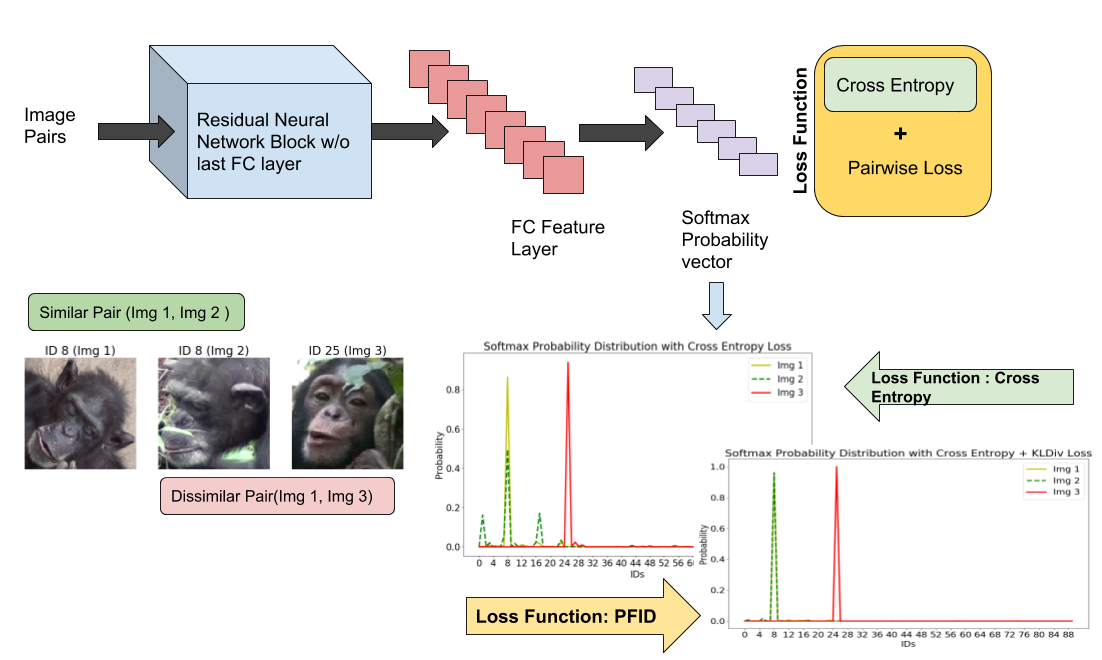}
  	\caption{Illustration of proposed PFID loss function vs. the standard cross entropy loss on the learned class probability distributions with ResNet model. }
		\label{fig: intro_example}
        \vspace{-0.5cm}	
\end{figure}
We now present our proposed PFID loss function for unique identification of primates using cropped facial images that can be obtained using state of the art deep learning based detectors. We note that images will be largely collected by the general public, professional monkey catchers and field biologists. Typically, we expect the images to be captured in uncontrolled outdoor scenarios, leading to significant variations in facial pose and lighting. These conditions are challenging for robust eye and nose detection, which need to be accurate in order to be useful for facial alignment. Consequently, we train our identification model to work without facial alignment and capture the semantic similarities of the underlying space. 

The proposed loss formulation combines the standard cross entropy network with a guided pairwise KL divergence loss imposed on similar and dissimilar pairs. Using pairwise loss terms ensure that the underlying features are more discriminative and generalize better. Our analysis in Sec. \ref{sec: cluster} show empirical evidence that the learned features are more clusterable than when trained with the standard cross-entropy loss. 

An illustration of the effect of loss function is shown in Figure \ref{fig: intro_example}. A similar pair corresponds to images of same individual, while a dissimilar pair corresponds to images from two different identities. The learned class probability distribution for a similar pair and dissimilar pair using two different loss functions is shown. In case of network trained with PFID loss, the class probabilities are maximally similar for a similar pair as oppose to standard cross entropy loss.

Let,  $ \calX =\{x_1,x_2,\ldots,x_n\}$ be the training dataset of $ n $ samples with $ l_i \in \{1,2,\ldots,K \}$ as the associated labels. We use the labeled training data to create sets of similar image pairs, $\calC_s\!=\!\{\!(i,\! j)\!:\! x_i,\! x_j\! \in\! \calX, l_i\! =\! l_j\}$, and that of dissimilar pair, $\calC_d\!=\!\{(i, j)\!:\!  x_i,\! x_j\! \in\! \calX, \!l_i\! \neq\! l_j\!\}$ for $ i,j\!\in\!\{1,2,\!\cdots\!,n\} $.
The KL divergence between two distribution $p^i$ and $q^j$ corresponding to points $x_i$ and $x_j$ is given by
\begin{align}
\label{eq:KL}
KL(p^{i}||q^{j}) =\sum_{k=1}^{K}{p_k^i \log\frac{p_k^i}{q_k^j}}
\end{align}
For a similar pair $ (i,j)\in\calC_s $, we use the symmetric variant of (\ref{eq:KL}) given by 
\begin{align}\label{eq: sim}
\calL_{s}^{ij} = KL(p^{i}||q^{j})+KL(q^{j}||p^{i})
\end{align}
and for a dissimilar pair $ (i,j)\in\calC_d $, we use its large-margin variant for improving discriminative power 
\begin{align}\label{eq: dsim}
\calL_{d}^{ij}\!=\! \max(0,m\!-\!KL(p^i||q^j))\! +\! \max(0,m\!-\!KL(q^j||p^i))
\end{align}
where $ m $ is the desired margin width between dissimilar pairs. It is important to note that during training, when both $ x_i $ and $ x_j $ are misclassified by the model, minimizing (\ref{eq: sim}) may lead to an increase in the bias.
\paragraph{\textbf{Guided Pairwise Loss Function}}
Since we use class labels for the cross-entropy loss, we incorporate them in the pairwise loss terms to guide the training. Subsequently, we modify the terms in (\ref{eq: sim}) and (\ref{eq: dsim})  to get the following guided KL divergence loss term 
\begin{align}
\calL_{s} = \sum_{i,j \in \calC_s}a \calL_{s}^{ij}, \qquad
\calL_{d} = \sum_{i,j \in \calC_d}a \calL_{d}^{ij}
\end{align}
where, $a= 1$ if either $ \arg \max \; p^i= l_i $ or $\arg \max \;  q^j = l_j$ and $a= 0$ otherwise. 
The loss function for PFID is given by the sum of standard cross entropy ($ \calL_{CE} $) and the guided KL divergence loss
\begin{align}
\calL(\theta) = \calL_{CE} + \frac{1}{|\calC_s|} \sum_{j,k \in \calC_s}{a\calL_{s}^{jk}} + \frac{1}{|\calC_d|} \sum_{j,k \in \calC_d}{a \calL_{d}^{jk}}
\end{align}
This loss function is used to train the network with a mini-batch gradient descent. Here $ |\calC_s| $ and $ |\calC_d| $ are the number of similar and dissimilar pairs respectively in a given batch. More details on the training are provided in Sec. \ref{sec:training_details}.


\section{Experimental Setup and Results}
\subsection{Dataset}
We evaluate our model using three datasets, the details of which are given in Table \ref{tab: datasets}. As is typical of wildlife data collected in uncontrolled environments, all the three datasets have a significant class imbalance as reported in the Table \ref{tab: datasets}.
\vspace{-0.2cm}
\subsubsection{Rhesus Macaque Dataset} 
\begin{figure}[h]
\vspace{-0.3cm}	
	\centering
			{\includegraphics[width =0.18\textwidth,height=1.5cm,keepaspectratio]{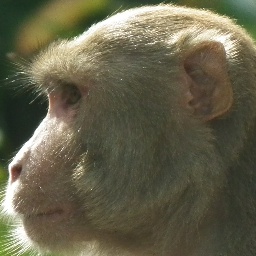}
        \includegraphics[width  =0.18\textwidth,height=1.5cm,keepaspectratio]{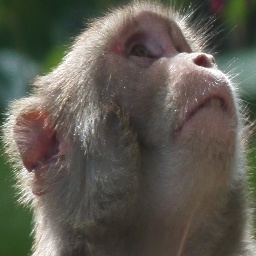}
       \includegraphics[width =0.18\textwidth,height=1.5cm,keepaspectratio]{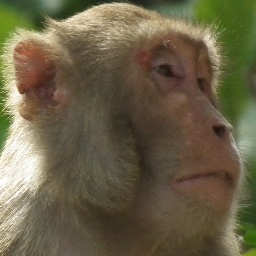}
       \includegraphics[width  =0.18\textwidth,height=1.5cm,keepaspectratio]{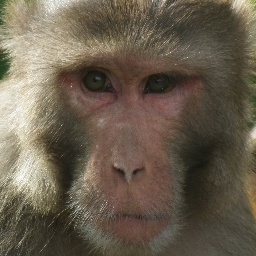}
        \includegraphics[width  =0.18\textwidth,height=1.5cm,keepaspectratio]{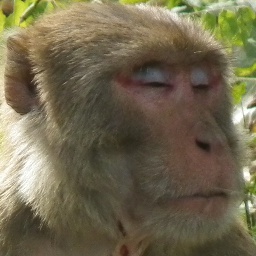}
        }\\
        {\includegraphics[width =0.18\textwidth,height=1.5cm,keepaspectratio]{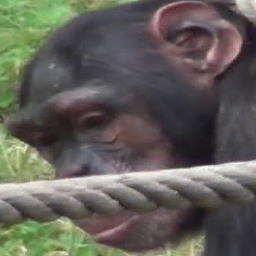}
        \includegraphics[width  =0.18\textwidth,height=1.5cm,keepaspectratio]{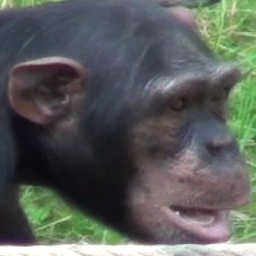}
       \includegraphics[width =0.18\textwidth,height=1.5cm,keepaspectratio]{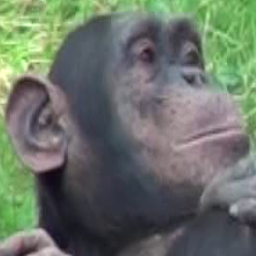}
       \includegraphics[width  =0.18\textwidth,height=1.5cm,keepaspectratio]{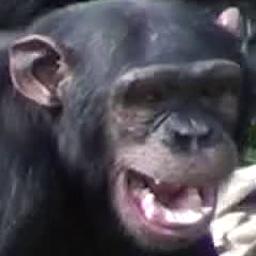}
        \includegraphics[width  =0.18\textwidth,height=1.5cm,keepaspectratio]{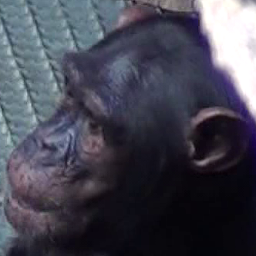}
        }
		\caption{{Pose variations for one of the Rhesus Macaque (Top) and Chimpanzee (Below) from the dataset.}}
		\label{fig: monk_example}
\end{figure}

The dataset is collected using DSLRs in their natural dwelling in an urban region in the state of Uttarakhand in northern India. The dataset is cleaned manually to remove images with no or very little facial content (e.g., extreme poses with only one ear or only back of head visible). The filtered dataset has 59 identities with a total of 1399 images. An illustrative set of pose variations for the datasets are shown using the cropped images in figure \ref{fig: monk_example}. Due to the small size of this dataset, we combined our dataset with the publicly available dataset by Witham \cite{witham2017automated}. The combined dataset comprises 7679 images of 93 individuals. 
Note that we use the combined dataset only for the individual identification experiments, as the public data by Witham comprises of pre-cropped images. On the other hand, the detection and the complete PFID pipeline is evaluated on a test set comprising full images from our macaque dataset.
\vspace{-0.5cm}
\subsubsection{Chimpanzee Dataset}
The C-Zoo and C-Tai dataset consists of 24 and 66 individuals with 2109 and 5057 images respectively \cite{freytag2016chimpanzee}. The C-Zoo dataset contains good quality images of chimpanzees taken in a Zoo, while the C-Tai dataset contains more challenging images taken under uncontrolled settings of a national park. We combine these two datasets to get 90 identities with a total of 7166 images.
\begin{table}[h!]
	\centering
		\begin{tabular}{|cccc|}
		\hline
  Dataset &  Rhesus Macaques &  C-Zoo& C-Tai  \\ \hline
 \# Samples &7679 &2109 &  5057  \\
 \# Classes & 93 & 24&66  \\
 \# Samples/individual & [4,192]& [62,111]& [4,416]\\\hline
        	\end{tabular}
	\caption{Dataset Summary. The numbers in the brackets show the range of samples per individual ([min,max]), highlighting the imbalance in the datasets.}
	\label{tab: datasets}
 \end{table}

\vspace{-0.3cm}
\subsection{Evaluation Protocol}
We evaluate and compare the performance of our PFID system under four different experimental settings, namely: classification, closed-set identification, open-set identification and verification. \\
\noindent\textbf{Classification}
To evaluate the classification performance the dataset is divided into $80/20$ train/test splits. We present the mean and standard deviation of classification accuracy over five stratified splits of the data. As opposed to other evaluation protocols discussed below, all the identities are seen during the training, with unseen samples of same identities in the test set. \\
\noindent\textbf{Open and Closed-Set Identification}
Both, closed-set and open-set performance is reported on \emph{unseen} identities. We perform 80/20 split of data w.r.t. to identities, which leads to a test set with 18 identities in test for both chimpanzee and macaque datasets. We again use five stratified splits of the data. For each split, we further perform 100 random trials for generating the probe and gallery sets. However, the composition of the probe and gallery sets for the closed-set scenario is different from that of open set. \\
\noindent\textbf{Closed-Set}: In case of closed-set identification, all identities of images present in the probe set are also present in the gallery set. Each probe image is assigned the identity that yields the maximum similarity score over the entire gallery set. We report the fraction of correctly identified individuals at Rank-1 to evaluate the performance. \\
\noindent\textbf{Open-Set}: In case of open-set identification, some of the identities in the probe set may not be present in the gallery set. This allows to evaluate the recognition system to validate the presence or absence of an identity in the gallery. To validate the performance, from the test of 18 identities, we used all the images of odd numbered identities as probe images with no images in the gallery. The rest of the even numbered identities are partitioned in the same way as closed-set identification to create probe and gallery sets. We report Detection and Identification Rate (DIR) at $1\%$ FAR to evaluate open-set performance.
\\\\
\noindent\textbf{Verification}
We compute positive and negative scores for each sample in test set. The positive score is the maximum similarity score of the same class and negative scores are the maximum scores from each of the classes except the true class of the sample. In our case, where the test data has 18 identities, each sample is associated with a set of 18 scores, with one positive score from the same identity and 17 negative scores corresponding to remaining 17 identities. The verification accuracy is reported as mean and standard deviation at $1\%$ False Acceptance Rates (FARs).
\subsection{Network Details and Parameter Setting}\label{sec:training_details}
We resize all the face images in macaque and chimpanzee dataset to $112 \times 112$. We add the following data augmentations: random horizontal flips and random rotations within $5$ degrees for both the datasets. We use the following base network architectures for PFID: ResNet-18 \cite{ResNet18_2016} and DenseNet-121 \cite{DenseNet} and remove the first maxpool layer because of small image size. For CE setting, we fine-tuned the imagenet pre-trained networks with cross-entropy loss and a batch size of 16. For the PFID setting, for each image in a batch, a similar class image is sampled to make a batch size of 8 pairs (16 images in a batch). The dissimilar pairs are then exhaustively created from these pairs. We used SGD for optimization with an initial learning rate of $10^{-3}$ and weight decay of $5e-4$. We trained all the models for both datasets for 40 epochs with learning rate decay by 0.1 at $25^{th}$ and $35^{th}$ epoch. We observed better performance with batch size of 16 instead of 32 or higher especially in case of training with only cross-entropy loss. It is recommended to use a lower batch size given that the training data is less in both the datasets.
\subsection{Results}
We present the results corresponding to PFID and other state of the art approaches for face recognition.
\vspace{-0.4cm}
\subsubsection{Baseline Results}
For the baseline results, we extracted the penultimate (FC) layer features from both ResNet-18 and DenseNet-121 models. For all the evaluation protocols, the features are $l2$-normalized and in addition for classification, they are used to train a SVM (Support Vector Machine) classifier by performing a grid-search over the regularization parameter. The results are given in the first 2 rows of the Table \ref{tab:chimp_results} and \ref{tab:maca_results}. We directly used the features and did not perform PCA (Principal Component Analysis) to
reduce the number of feature dimensions because it had no impact on the performance in each evaluation.
\vspace{-0.4cm}
\subsubsection{Comparison with state of the art approaches}
We compare PFID with recent work PrimNet \cite{Jain_2018} that achieved state of the art performance on chimpanzee face dataset. While our approach outperforms PrimNet by a large margin, it is worth noting that our results are reported on non-aligned face images, that makes PFID better suited for the application of crowdsourced population monitoring by eliminating the need for manual annotations of fiducial landmarks. Since ResNet-18 and DenseNet-121 are pretrained on imagenet data, we additionally fined-tuned ArcFace \cite{deng2018arcface} and SphereFace \cite{liu2017sphereface} models that are pre-trained on human face images, specifically on CASIA \cite{yi2014learning} dataset. We use ResNet-50 as the backbone for ArcFace and 20-layer network for SphereFace, and use the parameters given in the respective papers. We observed best performance with batch size 32 in all the three methods. We used a learning rate of 0.1, 0.01 and 0.001 for PrimNet (trained from scratch), SphereFace and ArcFace respectively and weight decay as $5e-4$. We trained all the models for 30 epochs to avoid over-fitting with learning rate decay by 0.1 at 15$^{th}$ and 25$^{th}$ epoch. The results are reported in Table \ref{tab:chimp_results} and \ref{tab:maca_results} for both the datasets. The results highlight that the imagenet pre-trained models generalize well in our case where the training data is not huge. Further, it should be noted that the results reported for the three models ArcFace, SphereFace and PrimNet are also reported without face alignment as oppose to the results reported in the respective papers. While we report results with non-aligned face images, we would also like to point out that the performance dropped in all the approaches with aligned face images in case of chimpanzee dataset owing to loss of features in aligned faces.
\begin{table}[t!]
	\centering
	\begin{tabular}{|c|c|c|c|c|}\hline
		\multirow{2}{*}{Method} & Classification& Closed-set & Open-set & Verification\\
		& Rank-1 & Rank-1& Rank-1& 1 \%  FAR  \\\hline 
		\hline
		Baseline (ResNet-18 FC + SVM) & 55.38 $\pm$ 1.18 & 70.51 $\pm$ 2.98 & 12.80 $\pm$ 5.73 & 37.10 $\pm$ 4.63  \\\hline
		Baseline (DenseNet-121 FC +SVM) &61.78 $\pm$ 1.4 &75.34 $\pm$ 3.98 & 30.51 $\pm$ 6.61 & 54.80 $\pm$ 3.65 \\\hline
		ArcFace (ResNet-50)   & 	85.47 $\pm$ 0.86 & 78.47 $\pm$ 5.81 & 41.24 $\pm$ 7.82 & 63.91  $\pm$ 5.37
			\\\hline 
		SphereFace-20   & 78.38 $\pm$ 1.23 & 72.72 $\pm$ 3.44 & 35.49 $\pm$ 8.34  & 57.74 $\pm$ 6.38\\\hline 
		PrimNet   &	
					 70.86 $\pm$ 1.19 & 72.22 $\pm$ 5.33  &  37.27 $\pm$ 5.48 & 62.83 $\pm$ 5.98 \\\hline 
		CE (ResNet-18)  & 	
			85.29 $\pm$1.43 & 86.44 $\pm$ 5.42 & 48.62 $\pm$ 9.05 & 75.19 $\pm$ 8.16 \\\hline 
		CE (DenseNet-121)  & 
			86.74 $\pm$ 0.74 & 87.01$ \pm$ 5.39 & 53.60 $\pm$ 13.04 & 76.86 $\pm$ 9.55  \\\hline 
		\textbf{PFID} (ResNet-18) & 
			88.98 $\pm$ 0.26 & 88.26 $\pm$ 5.01 & 59.36 $\pm$ 9.12 & 80.06 $\pm$ 6.62 \\\hline 
		\textbf{PFID} (DenseNet-121) & 
			\textbf{ 90.78 $\pm$ 0.53} & \textbf{91.87 $\pm$ 2.92}& \textbf{66.24 $\pm$ 8.08}   &\textbf{ 83.23 $\pm$ 6.07}  \\\hline 
	\end{tabular}
	\caption{Evaluation of Chimpanzee dataset for  classification, closed-set, open-set and verification setting. Baseline results are reported by taking the penultimate layer features of the network and training a SVM for classification. For all the remaining settings the features are directly used for the evaluation protocol.}
	\label{tab:chimp_results}
	\vspace{-0.5cm}
\end{table}

\begin{table}[t!]
	\centering
	\begin{tabular}{|c|c|c|c|c|}\hline
		\multirow{2}{*}{Method} & Classification& Closed-set & Open-set & Verification \\
		& Rank-1 & Rank-1& Rank-1& 1 \%  FAR  \\\hline 
		\hline
		Baseline (ResNet-18 FC +SVM) &85.28 $\pm$ 0.25 &88.29 $\pm$ 2.95 & 50.09 $\pm$ 7.35 &66.98 $\pm$ 9.21\\\hline
		Baseline (DenseNet-121 FC +SVM)  & 88.3 $\pm$ 0.57&89.24 $\pm$ 3.63 & 53.93 $\pm$ 10.27 & 71.34 $\pm$ 8.88\\\hline
		ArcFace (ResNet-50)  & 98.23 $\pm$ 0.47 & 93.98 $\pm$ 2.99 & 67.07 $\pm$ 13.91 & 95.16 $\pm$ 1.56\\\hline 
		SphereFace-20  & 97.61 $\pm$ 0.74 & 93.41 $\pm$ 2.19 & 95.62 $\pm$ 12.21 & 93.18 $\pm$ 1.95\\\hline 
		PrimNet   & 97.11 $\pm$ 0.65 & 90.94 $\pm$ 2.54 & 65.98 $\pm$ 15.23 & 92.14 $\pm$ 2.82 \\\hline 
		CE (ResNet-18)  & 97.91 $\pm$ 0.58 & 95.94 $\pm$ 2.94 & 79.69 $\pm$ 8.12 & 96.35 $\pm$ 2.06 \\\hline 
		CE (DenseNet-121)  & 97.99 $\pm$ 0.69 & 96.24 $\pm$ 0.85 &  71.36 $\pm$ 10.05 & 96.01 $\pm$ 3.01 \\\hline 
	\textbf{PFID} (ResNet-18)  & 98.71 $\pm$ 0.41 & 96.18 $\pm$ 1.58 & 83.02 $\pm$ 7.36 & 97.71 $\pm$ 0.91\\\hline 
		\textbf{PFID} (DenseNet-121) &\textbf{98.91 $\pm$ 0.40} & \textbf{97.36 $\pm$ 1.73} & \textbf{84.00 $\pm$ 7.43} & \textbf{98.24 $\pm$ 0.94}  \\\hline 
	\end{tabular}
	\caption{Evaluation of Rhesus Macaque dataset for  classification, closed-set, open-set and verification setting. Baseline results are reported by taking the penultimate layer features of the network and training a SVM for classification. For all the remaining settings the features are directly used for the evaluation protocol.}
	\label{tab:maca_results}
\end{table}
\vspace{-0.5cm}
\subsubsection{PFID Results}
\begin{figure*}[h!]
	\centering
        { \includegraphics[width =0.45\textwidth,trim={0.5cm 0.5cm 0cm 0.5}]{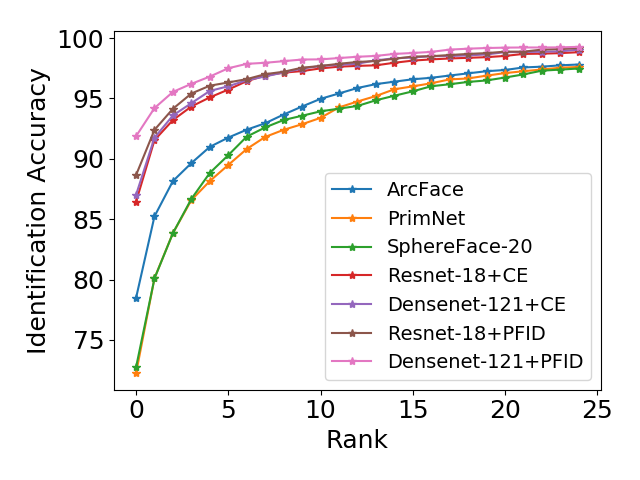}\includegraphics[width =0.45\textwidth,trim={0.5cm 0.5cm 0cm 0.5}]{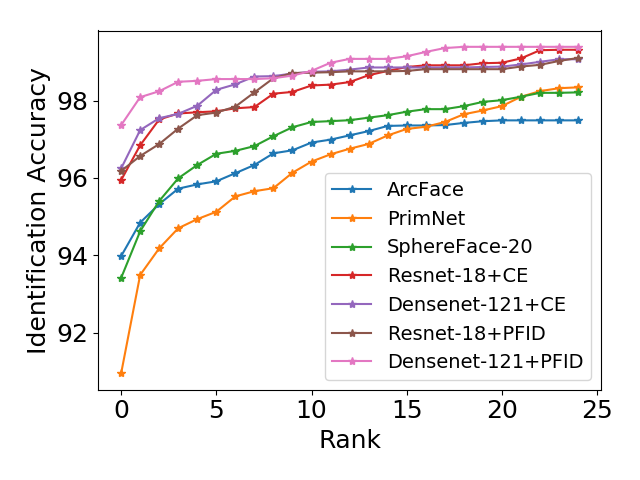}}
        \\
        { \includegraphics[width =0.45\textwidth,trim={0.5cm 0.5cm 0cm 0}]{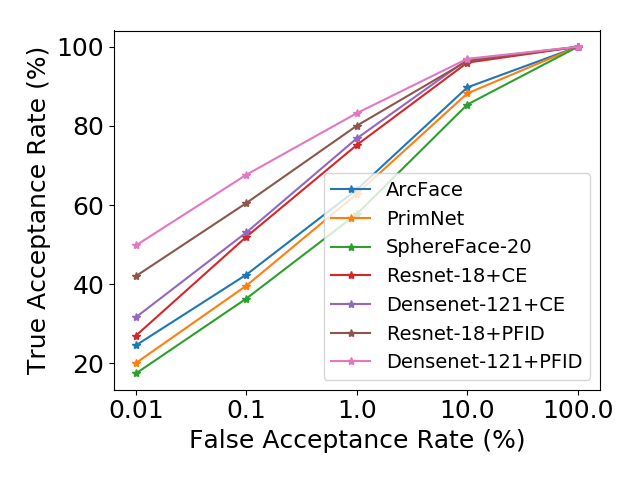}\includegraphics[width =0.45\textwidth,trim={0.5cm 0.5cm 0cm 0}]{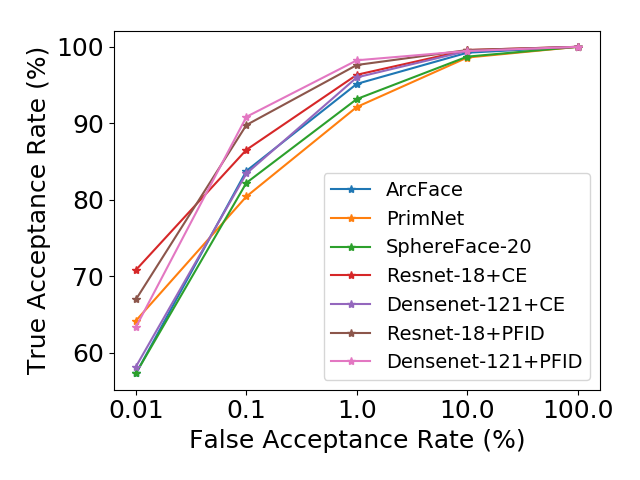}}
      \vspace{-0.3cm}	
		\caption{CMC (Top) and TAR vs FAR (Bottom) plots for (Left) C-Zoo+CTai and (Right) Rhesus Macaques dataset.}
		\label{fig: cmc_tar}    
\end{figure*}

To show the efficiency of our approach, we fine-tuned ResNet-18 and DenseNet-121 models with standard cross entropy (CE) loss and report in the Table \ref{tab:maca_results} and \ref{tab:chimp_results} for macaque and chimpanzee datasets respectively and compared it with the PFID loss. We observe an increase in performance for the four evaluation protocols with PFID loss as opposed to traditional cross entropy fine-tuned network. Imposing a KL-divergence loss has improved the discriminativeness of features by skewing the probability distributions of similar and dissimilar pairs. For chimpanzee dataset an improvement of $\textbf{4.04\%}$, $\textbf{4.86 \%}$, $\textbf{12.64\%}$ and $\textbf{6.97 \%}$ is achieved in case of classification, closed-set, open-set and verification respectively using DenseNet-121. The corresponding CMC (Cumulative Matching Characteristic) and TAR (True Acceptance Rate) vs FAR plots for the datasets are shown in Figure \ref{fig: cmc_tar}.
\vspace{-0.6cm}
\subsubsection{Feature Learning and Generalization}\label{sec: cluster} To further show the effectiveness of PFID loss function and robustness of features, we perform cross dataset experiments in Table \ref{tab:transfer}. We used model trained on chimpanzee dataset and extracted features on macaque dataset to evaluate the performance for closed-set, open-set and verification task and vice versa. We compared the quality of the features learned with PFID with the features learned with cross entropy based fine-tuning. We also show the generalizability between two chimpanzee datasets captured in different environments i.e. CZoo and CTai. The results clearly highlight the advantage of PFID over cross entropy loss for across data generalization. Additionally, to highlight the discriminativeness and clusterability of the class specific features, we cluster the feature representations of unseen (identities) test data using K-means clustering algorithm. We report the clustering performance in Table \ref{tab:cluster} and compare with the standard cross entropy loss. 

\vspace{-0.3cm}
\begin{table}
{\small 
\centering
		\begin{tabular}{|l|c||c||}
		\hline
      Model & \multicolumn{1}{c|}{Macaque}& \multicolumn{1}{c|}{Chimpanzee}\\\cline{2-3}
       & NMI &  NMI \\\hline
       CE & 0.868 $\pm$ 0.008 & 0.686 $\pm$ 0.084 \\
       \hline
        PFID & 0.897 $\pm$ 0.030 & 0.715 $\pm$ 0.089 \\
       \hline
\end{tabular}
\caption{Comparison of K-means clustering performance on the learned representations with DenseNet-121. The results highlight that the PFID learns more clusterable space.}
\label{tab:cluster}
}
\end{table}

\vspace{-1.2cm}
\begin{table}[h]
\centering
		\begin{tabular}{|c|c|c|c|c|c|c|c|c|}
		\hline
       & \multicolumn{2}{c|}{Macq. $\rightarrow$ Chimp.}& \multicolumn{2}{c|}{Chimp. $\rightarrow$ Macq.}& \multicolumn{2}{c|}{CZoo $\rightarrow$ CTai} & \multicolumn{2}{c|}{CTai $\rightarrow$ CZoo} \\\cline{2-9}
       & CE & PFID & CE & PFID & CE & PFID & CE & PFID\\\hline
       Closed Set & 54.58 & \textbf{63.48} & 83.02 & \textbf{88.38} & 59.92 & \textbf{70.35} & 87.54 & \textbf{91.96}  \\
       \hline
       Open Set & 13.56 & \textbf{34.29} & 32.04 & \textbf{43.00} & 17.21 & \textbf{27.21} & 43.25 & \textbf{64.75} \\
       \hline
       Verification & 43.02 & \textbf{63.77} & 67.51 & \textbf{75.37} & 48.68 & \textbf{60.57} & 66.71 & \textbf{82.22}  \\
       \hline
\end{tabular}
\caption{Evaluation of learned model across datasets. Left of the arrow indicates the dataset on which the model was trained on, and right of the arrow indicates the evaluation dataset. All the results are reported for DenseNet-121 network.}
\label{tab:transfer}
 \end{table}
 \vspace{-0.5cm}
        
\subsubsection{Comparison with Siamese Network based features} One might draw similarity of our approach with the popular siamese networks \cite{Schroff_2015_CVPR} that are trained on similar and dissimilar pairs to result in a similairty score at the output. We train ResNet-18 on chimpanzee data in siamese setting with pairwise hinge-loss on features to show that the learned features in the classification setting are not discriminative as compared to our PFID. While training in siamese setting, we also observe that the network overfits on the training data and performs poorly on unseen classes. The results for different evaluation protocols are: Classification (83.97 $\pm$ 1.42), Closed-set (75.45 $\pm$ 5.51), Verification (57.28 $\pm$ 7.37) and Open-set (22.22 $\pm$ 8.07).
\vspace{-0.4cm}
\subsubsection{Identification on Detected Face Images} The above results evaluated the performance of PFID on cropped face images \emph{i.e.} the true bounding box of the test samples.  As the captured images with handheld devices like cameras would also have background,  we evaluate the performance of PFID on the detected faces on test samples. Since, we had 1191 full images for the Macaque dataset,  the detector is trained and tested with a split of 80/20. We fine-tune state-of-the-art Faster-RCNN \cite{Ren_NIPS_FRCNN2015} detector for detecting macaque faces and achieve highly accurate face detection performance. The identification results on the cropped faces obtained from the detector is shown in Table \ref{tab: detected_maca_results}. For identification evaluation, we have 10 identities and 227 images for both closed-set and verification, whereas for open-set we extend the probe set by adding 8 identities and 1100 samples which are not part of the dataset. 
\vspace{-0.4cm}
\subsection{Integration with Crowd Sourcing App}
We have developed a simple app to work as a front-end for PFID, which permits a user to upload geo-tagged images of individuals and troops as well as report a conflict incident. Augmented with the PFID based back-end service, this app could help maintain an updated database of reported conflicts, along with a primate database indexed by individuals, troop and last-sighted locations, which can be used with techniques like Capture-Recapture to estimate population densities. 
\vspace{-0.3cm}
\begin{table}
{\small 
\centering
	{\fontsize{8}{9}\selectfont
		\begin{tabular}{|l|c|c|c|}
		\hline
        Method & Closed-set & Open-set & Verification \\
        & Rank-1& Rank-1& 1 \%  FAR \\\hline 
       CE (ResNet-18) & 95.00 & 70.78 & 89.22 \\\hline
       PFID (ResNet-18) & 97.20 & 78.80 & 91.11 \\\hline
       CE (DenseNet-121) & 95.30 & 80.67  & 91.56  \\\hline
       PFID (DenseNet-121) & \textbf{97.80} & \textbf{89.67} & \textbf{95.11} \\\hline
       \end{tabular}
       }
        \caption{Evaluation of detected macaque faces for closed set, open set and verification setting.}
        \label{tab: detected_maca_results}
        }
\vspace{-0.4cm}
\end{table}

\vspace{-0.3cm}
\vspace{-0.5cm}
\section{Conclusion}
In this work, we discussed the problem of unique identification of non-human primates using face images captured in the wild. From existing literature, we found that population monitoring is an important step in the management strategies and largely rely only on field-based efforts. In this work, we identified this challenge and proposed an alternate solution that can simultaneously improve monitoring of commensal primates as well as actively involve the affected human community without any serious cost implications. We developed a novel face identification approach that is capable of learning pose invariant features, thus allowing to generalize well across poses without the requirement of a face alignment step. Additionally, the proposed approach leverages the pairwise constraints to capture underlying data semantics enabling it to perform effectively for unseen classes.  With the effectiveness of our approach in different identification tasks on real world data, we foresee that the PFID system could become a part of widely used wildlife management tools like SMART\footnote{http://smartconservationtools.org/}.

\section*{Acknowledgement}
\vspace{-0.5cm}
This work is supported by Microsoft AI for Earth Grant 2017-18 and Infosys Center for AI at IIIT Delhi, India.
\vspace{-0.5cm}

{\small
\bibliographystyle{splncs04}
\bibliography{Monkey_main}
}
%
%
%




\end{document}